# Persistent Monitoring of Stochastic Spatio-temporal Phenomena with a Small Team of Robots

Sahil Garg and Nora Ayanian
Department of Computer Science, University of Southern California, Los Angeles, California 90089
Email: {sahilgar, ayanian}@usc.edu

*Abstract*—This paper presents a solution for persistent monitoring of real-world stochastic phenomena, where the underlying covariance structure changes sharply across time, using a small number of mobile robot sensors. We propose an adaptive solution for the problem where stochastic real-world dynamics are modeled as a Gaussian Process (GP). The belief on the underlying covariance structure is learned from recently observed dynamics as a Gaussian Mixture (GM) in the low-dimensional hyper-parameters space of the GP and adapted across time using Sequential Monte Carlo methods. Each robot samples a belief point from the GM and locally optimizes a set of informative regions by greedy maximization of the submodular entropy function. The key contributions of this paper are threefold: adapting the belief on the covariance using Markov Chain Monte Carlo (MCMC) sampling such that particles survive even under sharp covariance changes across time; exploiting the belief to transform the problem of entropy maximization into a decentralized one; and developing an approximation algorithm to maximize entropy on a set of informative regions in the continuous space. We illustrate the application of the proposed solution through extensive simulations using an artificial dataset and multiple real datasets from fixed sensor deployments, and compare it to three competing state-of-the-art approaches.

## I. Introduction

In scenarios such as natural disasters, seasonal agriculture, and other short-duration operations, a rapidly deployable, autonomous mobile sensing system that decides where to take sensor measurements can be more versatile and cost-effective than installing stationary sensors. In this work, we are interested in formulating a solution for persistent sensing of real-world stochastic phenomena using a team of mobile robots, even when the *underlying covariance structure changes sharply across time*, such as sunlight variation in a forest understory (Fig. 1). Assuming no prior knowledge on the underlying model of the phenomenon dynamics, this presents two challenges: 1) adapting a belief on the underlying model based on recently observed phenomenon dynamics and 2) correspondingly optimizing the next sensing locations.

While exactly modeling stochastic real-world phenomena remains a significant challenge, this work deals mainly with modeling the underlying covariance structure. The underlying covariance structure directly corresponds to information metrics such as entropy, required for evaluating the informativeness or representativeness of sensor readings across a set of locations [9, 13, 30]. Gaussian processes (GP) have emerged as a favored choice for this specific modeling goal primarily because of their nonparametric nature [14, 20, 33, 42]. Modeling stochastic phenomena as GPs allows the flexibility of representing the underlying global covariance structure with the small set of parameters of a covariance function (termed *hyper-parameters*). This motivates us to model a real-world phenomenon as a GP for our problem.

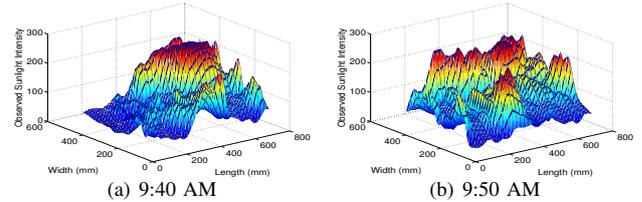

(a) 9:40 AM  (b) 9:50 AM

Fig. 1: Sunlight dynamics across space under a tree canopy in a forest reserve. The figure demonstrates high spatio-temporal dynamics, underlying spatial covariance structure changes across time, and the multi-scale nonstationary nature of the phenomenon.

With this simplification, the two challenges become: 1) continually adapting belief on the covariance structure in the low-dimensional hyper-parameter space based on observations in the previous sensing cycle; and 2) maximizing entropy on the set of locations chosen for sensing in the next sensing cycle. We propose a system for addressing these challenges in a separable yet unified manner, where the belief adaptation occurs on a central server (this may be a robot), and the entropy maximization is decentralized across the robot team.

Briefly, our approach is as follows. The covariance structure belief initiates randomly on the central server. Each time a robot communicates data to the server, the belief is updated and sent to that robot. The robot uses a sample from the belief to independently optimize the next set of informative sensing locations (entropy maximization). While the server has a single belief adapted over time, since the robots communicate asynchronously, at any given time each robot is operating on a different belief, adapted with the data available when that robot last successfully communicated to the server.

We represent the belief as a Gaussian Mixture (GM) [3, 37, 41] in the hyper-parameter space that is adapted after every sensing cycle using a Sequential Monte Carlo (particle filtering) technique [1, 7, 16, 26, 29, 35]. Then, each robot optimizes its set of sensing locations in the next cycle such that the entropy function (evaluated using a sample from the GM-based belief) attains its maximum value on the optimal set of locations. This way, each robot senses the phenomenon as a single-scale stochastic process, while the joint decentralized sensing effort of the team leads to multi-scale sensing [2, 5, 17, 36, 48]. Since many real world phenomena exhibit multi-scale stochastic dynamics, the problem we address has many applications [44]. We will show that the proposed approach proves to be effective in the extensive empirical validation presented in this paper.

Maximizing entropy on a set of locations is known to be NP-hard even for discrete cases. Since entropy is a submodular function, nearly-optimal greedy solutions exist with polynomial cost [33]. We propose an approximation algorithm that greedily learns a set of informative regions in the continuous 2-D space. The intuition behind optimizing a region instead of a location is that there are infinitely many points in the continuous space with approximately equal entropy. We approximate sensing across a region by sensing at a sparse set of randomly sampled locations from that region. This flexibility of random sensing in a region can be further exploited for planning decentralized obstacle-free trajectories to the regions [18].

*A. Contributions*

The key contributions of this work are three-fold. First, we present an approach for adapting belief on the underlying covariance structure of a phenomenon using *particle filtering supplemented by Markov Chain Monte Carlo (MCMC) sampling*. In current approaches using only particle filtering, a sharp change in the covariance structure prevents survival of the belief particles. Here, the belief particles distribution is compactly represented as a GM and the efficient adaptation of the GM-based belief is ensured by using MCMC sampling if the conditional entropy on the GM distribution is high. Second, we *exploit the belief on the covariance structure to transform the problem of entropy maximization* on the dynamics distribution observed by multiple robots into multiple decentralized optimizations where each robot maximizes entropy on the dynamics distribution for its own sensing. Third, since the entropy maximization problem is NP-hard even for the decentralized case, we develop an approximation algorithm that *maximizes entropy on a set of informative spatial regions* in the continuous space using MCMC sampling.

*B. Related Work*

Many variants of this problem have been previously studied. The problem of tracking a nonlinear non-Gaussian target is solved by maximizing mutual information between the joint observation and target state distributions [24, 27, 28]. In these works, the primary objective is tracking the target state, while the present work tracks the state of the underlying covariance structure for optimizing persistent sensing of the stochastic dynamics. Smith et al. [46] persistently monitor a fixed accumulation field using a linear programming based formulation. Others consider a time-invariant sensory function directly representing the importance of a spatial location [12, 43, 47], in contrast to the complex spatio-temporal stochastic field sensing considered here.

Another class of work addresses informative sensing of Gaussian process dynamics while assuming the covariance structure known and static [11, 22, 25, 34, 45]. Singh et al. [45] propose a Sequential Allocation (SA) based centralized strategy that extends an algorithm meant to optimize a single informative path to the case of multiple paths. Le Ny and Pappas [34] nonmyopically optimize an informative trajectory for a robot in the discrete space using Gaussian Markov Random Field modeling [4, 31]. Graham and Cortés [22] exploit near-independence between distinct sensing locations for a generalized Voronoi partition based solution. For a cost budget and motion constraints, Hollinger and Sukhatme [25] sample informative trajectories in the continuous space using a branch and bound technique. Chen et al. [11] formulate a broadcast communication-based distributed solution for selecting the next discrete sensing location for each robot.

Contrary to the related works that solve only part of our problem, this work formulates a system solution for the problem in a unified manner.

## II. PROBLEM FORMULATION

Consider persistent sensing of stochastic dynamics in a continuous spatial region $\boldsymbol{R} \subset \mathbb{R}^2$ (nonconvex region). In this context, it suffices to sense across a sparse set of locations due to correlation between observations in the same neighborhood. It may also be infeasible for a robot to take accurate sensor readings while moving. For instance, if a robot senses water temperature while in motion, the reading would be inaccurate due to the temporal disturbance caused by motion in the neighborhood of the robot. For some phenomena (e.g. fluorescence) sensing is costly even if accurate while moving; such scenarios benefit significantly from algorithms that enable sensing at a sparse set of informative locations, which is the focus of this work. Regardless of cost, it is not feasible to physically sense the dynamics across the entire continuous space.

Consider a team of $r$ mobile robots $\{1, \cdots, r\}$ that observe $\boldsymbol{y}^{(1 \cdots r)} = \{\boldsymbol{y}^{(1)} \in \mathbb{R}^{n_1}, \ldots, \boldsymbol{y}^{(r)} \in \mathbb{R}^{n_r}\} \in \mathbb{R}^n$ across sets of spatial sensing locations $\boldsymbol{X}^{(1 \cdots r)} = \{\boldsymbol{X}^{(1)}, \ldots, \boldsymbol{X}^{(r)} \subset \boldsymbol{R}\}$ at time $\boldsymbol{t}^{(1 \cdots r)} = \{\boldsymbol{t}^{(1)}, \cdots, \boldsymbol{t}^{(r)}\}$ respectively in each sensing cycle. The duration of a sensing cycle is not fixed and indirectly depends upon the predetermined number of sensing locations, motion capabilities of the robot, and practical issues such as delays due to communication failures, obstacle avoidance, etc. For persistent monitoring, this sensing cycle is repeated.

For each sensing cycle, the locations $\boldsymbol{X}^{(1 \cdots r)}$ are optimized in terms of *informativeness* [13]; the corresponding times $\boldsymbol{t}^{(1 \cdots r)}$ are not optimized since it may not be reasonable for a robot to traverse and sense across the locations with hard time constraints. To optimize the locations' informativeness, we model the dynamics as a spatial stochastic process where the spatial stochastic model is adapted after every sensing cycle. The spatial stochastic dynamics distribution across locations $\boldsymbol{X}^{(1 \cdots r)}$ is denoted as $\boldsymbol{\mathcal{Y}}^{(1 \cdots r)}$. Applying information theory, locations $\boldsymbol{X}^{(1 \cdots r)}$ are optimized s.t. entropy on $\boldsymbol{\mathcal{Y}}^{(1 \cdots r)}$ is maximized; i.e. minimizing conditional entropy across the unobserved region $\boldsymbol{R} \setminus \boldsymbol{X}^{(1 \cdots r)}$ [33]:

$$\boldsymbol{X}_*^{(1 \cdots r)} = \mathrm{argmax}_{\boldsymbol{X}: \boldsymbol{X} \subset \boldsymbol{R}} \mathcal{H}(\boldsymbol{\mathcal{Y}}), \tag{1}$$

where $\mathcal{H}(.)$ is the entropy function. Henceforth, for clarity of presentation, on the r.h.s. of entropy maximization expressions, a set of locations under an optimization is simply denoted as $\boldsymbol{X}$ (instead of $\boldsymbol{X}^{(1 \cdots r)}$) and the corresponding dynamics distribution as $\mathcal{Y}$ (instead of $\mathcal{Y}^{(1\cdots r)}$); the actual locations can be determined from l.h.s. or from the context.

Since computing entropy on an arbitrary stochastic distribution is intractable, $\mathcal{Y}$ is typically assumed to be a joint Gaussian distribution

$$\mathcal{Y} = \mathcal{N}(\boldsymbol{\mu}, \boldsymbol{\Sigma}); P(\boldsymbol{y}|\boldsymbol{\mu}, \boldsymbol{\Sigma}) = e^{(-(\boldsymbol{y}-\boldsymbol{\mu})^{\mathrm{T}}\boldsymbol{\Sigma}^{-1}(\boldsymbol{y}-\boldsymbol{\mu})/2)}/(2\pi)^{\frac{n}{2}}|\boldsymbol{\Sigma}|$$

with $\boldsymbol{y} \in \mathbb{R}^n$, $\boldsymbol{\mu} \in \mathbb{R}^n$, and $\boldsymbol{\Sigma}_{n \times n}$ representing respectively observation samples, mean (typically taken as zero), and positive semi-definite covariance matrix [42]. For the joint Gaussian $\mathcal{Y}$, $\mathcal{H}(\mathcal{Y}|\boldsymbol{\Sigma}) = \frac{1}{2}\log((2\pi e)^n|\boldsymbol{\Sigma}|)$ [42, page 203] is computed in $O(n^3)$ time (cost for the determinant evaluation).

Gaussian processes that are an infinite dimensional extension of joint Gaussians define covariance $\boldsymbol{\Sigma}_{ij}$ between two locations $\boldsymbol{x}_i, \boldsymbol{x}_j$ using a covariance function $\mathcal{K}(\boldsymbol{x}_i, \boldsymbol{x}_j)$ such as the *squared exponential kernel*:

$$\mathcal{K}(\boldsymbol{x}_i, \boldsymbol{x}_j) = \sigma_f^2 e^{-(\boldsymbol{x}_i-\boldsymbol{x}_j)^T diag(\boldsymbol{\sigma_l}^2)^{-1}(\boldsymbol{x}_i-\boldsymbol{x}_j)/2} + \sigma_n^2 \delta(\boldsymbol{x}_i, \boldsymbol{x}_j).$$

The first and second term model covariance on white noise and observation noise, respectively, and the kernel parameters are collectively termed $\boldsymbol{\theta} = \{\sigma_f, \sigma_n, \boldsymbol{\sigma_l}\} \in \mathbb{R}^4$ (also the hyper-parameters of the GP). Therefore, if the dynamics are modeled as a GP, the problem of entropy evaluation on $\mathcal{Y}$ narrows to knowing $\boldsymbol{\theta}$. In reality, one can only have a *belief* on $\boldsymbol{\theta}$, $P(\boldsymbol{\theta})$.

We can now formally present our problem.

**Problem 1** (Sensing of a Stochastic Phenomenon). *For a set of $r$ robots operating in a region $\boldsymbol{R}$, and a prior belief $\mathcal{P}(\boldsymbol{\theta})$,*
  1) *optimize locations informativeness: find locations $\boldsymbol{X}^{(1\cdots r)} \subset \boldsymbol{R}$ by maximizing entropy on the joint Gaussian distribution $\mathcal{Y}^{(1\cdots r)}$ across the locations, i.e. optimizing informativeness of $\boldsymbol{X}^{(1\cdots r)}$;*
  2) *adapt belief: find posterior belief $\mathcal{P}(\boldsymbol{\theta}|\boldsymbol{y})$ as per the observations $\boldsymbol{y}^{(1\cdots r)}$ sensed across locations $\boldsymbol{X}^{(1\cdots r)}$ by $r$ robots.*

Since Problem 1 is intractable to solve for an arbitrary belief distribution, we propose an algorithmic formulation for solving the problem approximately in finite time (and exactly in infinite time).

## III. Decentralization of Entropy Optimization & Adaptation of Belief

A key idea of our approach for adapting the belief on the GP hyper-parameters ($\boldsymbol{\theta}$) is to use particle filtering [16] after every sensing round of a robot; this is performed on a central server which can maintain communication with the robots. In case of communication failure, a robot adapts its belief on $\boldsymbol{\theta}$ locally. No communication is required between robots.

Another key aspect is approximating an arbitrary random belief distribution $\mathcal{P}(\boldsymbol{\theta})$ as a Mixture of Gaussians (GM), $\mathcal{GM}_{\boldsymbol{\theta}}$ [4, page 110]. This continuous parametric representation (GM) of the belief distribution enables: 1) low bandwidth communication of the belief between server and robot; 2) approximate yet efficient estimation of entropy on the belief distribution (for belief adaptation in Problem 1); 3) decentralized sampling of one belief point by each robot which jointly act as a sparse representation of the entire belief distribution (for decentralized optimization of sensing locations in Problem 1). On these lines, prior $\mathcal{P}(\boldsymbol{\theta}) = \mathcal{GM}_{\boldsymbol{\theta}}$ is initialized with $k$ Gaussian components at line 2 in Algorithm 1.

**Problem 2** (Joint Entropy Maximization). *For a set of $r$ robots $\{1, \cdots, r\}$ and a belief $\mathcal{GM}_{\boldsymbol{\theta}}$, optimize the sets of locations $\boldsymbol{X}_*^{(1\cdots r)} = \{\boldsymbol{X}_*^{(1)}, \cdots, \boldsymbol{X}_*^{(r)} \subset \boldsymbol{R}\}$ for the sensing s.t. $\boldsymbol{X}_*^{(1\cdots r)} = \mathrm{argmax}_{\boldsymbol{X}:\boldsymbol{X}\subset\boldsymbol{R}}\mathcal{H}(\mathcal{Y}|\mathcal{GM}_{\boldsymbol{\theta}})$*

In Problem 2, since it is intractable to evaluate $\mathcal{H}(\mathcal{Y}|\mathcal{GM}_{\boldsymbol{\theta}})$, one can approximate using samples $\boldsymbol{\theta}^{[1\cdots p]} \sim \mathcal{GM}_{\boldsymbol{\theta}}$:

$$\lim_{p\to\infty} \boldsymbol{X}_*^{(1\cdots r)} = \mathrm{argmax}_{\boldsymbol{X}:\boldsymbol{X}\subset\boldsymbol{R}}\mathcal{H}(\mathcal{Y}|\boldsymbol{\theta}^{[1\cdots p]}); \quad (2)$$

$$\lim_{p\to\infty} \mathcal{H}(\mathcal{Y}|\mathcal{GM}_{\boldsymbol{\theta}}) = \mathcal{H}(\mathcal{Y}|\boldsymbol{\theta}^{[1\cdots p]}) = \sum_{i=1}^{p}\mathcal{H}\left(\mathcal{Y}|\boldsymbol{\theta}^{[i]}\right) \quad (3)$$

In this centralized optimization (2), computing $\mathcal{H}(\mathcal{Y}|\boldsymbol{\theta}^{[1\cdots p]})$ is expensive for $p \gg 1$. Additionally, an efficient algorithm would be required to optimize the assignment of $\boldsymbol{X}_*^{(1\cdots r)}$ among the $r$ robots [45]. Despite a high computational cost, empirical evaluations do not provide promising results (see results for SA-GMTA in Table I(c)). Thus, we transform the above optimization into multiple decentralized optimizations:

$$\max_{\boldsymbol{X}:\boldsymbol{X}\subset\boldsymbol{R}}\sum_{i=1}^{r}\mathcal{H}\left(\mathcal{Y}|\boldsymbol{\theta}^{(i)}\right) \leq \sum_{i=1}^{r}\max_{\boldsymbol{X}^{(i)}:\boldsymbol{X}^{(i)}\subset\boldsymbol{R}}\mathcal{H}\left(\mathcal{Y}^{(i)}|\boldsymbol{\theta}^{(i)}\right)$$

Thus, a decentralized version of Problem 2 is stated:

**Problem 3** (Decentralizing Entropy Maximization). *For each robot $r_{id} \in \{1, \cdots, r\}$ sharing a belief $\mathcal{GM}_{\boldsymbol{\theta}}$,*
  1) *sample a belief point $\boldsymbol{\theta}^{(r_{id})} \sim \mathcal{GM}_{\boldsymbol{\theta}}$;*
  2) *optimize $\boldsymbol{X}_*^{(r_{id})} \subset \boldsymbol{R}$ for the sensing s.t. $\boldsymbol{X}_*^{(r_{id})} = \mathrm{argmax}_{\boldsymbol{X}:\boldsymbol{X}\subset\boldsymbol{R}}\mathcal{H}(\mathcal{Y}|\boldsymbol{\theta}^{(r_{id})})$.*

The subproblem of a single decentralized entropy maximization is separately addressed in Sec. IV. Thus, in Algorithm 1 line 3, all robots are notified to first optimize sensing locations locally using the belief $\mathcal{GM}_{\boldsymbol{\theta}}$, then make observations. Having the sensed data communicated to the server by robots, the next step is adapting the belief on the covariance structure.

**Problem 4** (Belief Adaptation). *For a prior belief $\mathcal{GM}_{\boldsymbol{\theta}}$ and joint data $\{\boldsymbol{X}^{(1\cdots r)}, \boldsymbol{y}^{(1\cdots r)}\}$ sensed by robots $\{1, \cdots, r\}$ in the previous sensing cycle, adapt the belief to $\overline{\mathcal{GM}}_{\boldsymbol{\theta}}^*$ on the central server s.t. $\overline{\mathcal{GM}}_{\boldsymbol{\theta}}^* = \mathrm{argmin}_{\overline{\mathcal{GM}}_{\boldsymbol{\theta}}}\mathcal{H}(\overline{\mathcal{GM}}_{\boldsymbol{\theta}}|\boldsymbol{y}^{(1\cdots r)})$.*

In Problem 4, belief is adapted as per the observations sensed by all robots. In Algorithm 1, however, the belief is adapted by the server using only the data $\{\boldsymbol{X}, \boldsymbol{y}\}$ sensed by a single robot[1] (line 5) so as to enable asynchronous communication between the server and the robots. The adaptation technique, presented next, is equally applicable to both cases.

The first step toward learning the adapted belief, $\overline{\mathcal{GM}}_{\boldsymbol{\theta}}$, is taking $p$ samples, $\boldsymbol{\theta}^{[1\cdots p]}$, from the prior belief, $\mathcal{GM}_{\boldsymbol{\theta}}$, and then evaluating the Gaussian likelihood, $P(\boldsymbol{y}|\boldsymbol{X}, \boldsymbol{\theta}^{[i]})$, as

---
[1] superscript "$(r_{id})$" is omitted from notations of datapoints sensed by a robot with id:$r_{id}$ to more clearly explain the belief adaptation technique

**Algorithm 1** Optimizing Persistent Sensing

1: Require: $r$, $p$, $spp$, $opp$, $k$
2: $\mathcal{GM}_\theta$ = initGM($k$); % initialize GM
3: **for** $r_{id} = 1 \to r$ **do** notifyRobotToSns( $r_{id}$, $\mathcal{GM}_\theta$ );
4: **while** persistent sensing continued **do**
5:    $\{X, y, t, r_{id}\}$ = waitFrRobotSns(); % wait for sensing
6:    **for** $i = 1 \to p$ **do** $\theta^{[i]} \sim \mathcal{GM}_\theta$; $w^{[i]} = P(y|X, \theta^{[i]})$;
7:    $\bar{w}^{[1 \cdots p]}$ = normalizeWeights($w^{[1 \cdots p]}$);
8:    $\mathcal{H}(\mathcal{GM}_\theta|y) = \log(1/p) - \sum_{i=1}^{p} \bar{w}^{[i]} \log\left(\bar{w}^{[i]} P(\theta^{[i]})\right)$;
9:    $\mathcal{H}(\overline{\mathcal{GM}}_\theta^*|y) = -\log(1/p)$; % entropy on optimal belief
10:    $epp = 100 * \exp\left(\mathcal{H}(\overline{\mathcal{GM}}_\theta^*|y) - \mathcal{H}(\mathcal{GM}_\theta|y)\right)$;
11:    **if** $epp < opp$ **then**
12:      **if** $epp < spp$ **then**
13:         $\{\theta^{[p+1\cdots 2p]}, w^{[p+1\cdots 2p]}\}$ = mcmcSample( $X$, $y$ );
14:         $\bar{w}^{[1\cdots 2p]}$ = normalizeWeights($w^{[1\cdots 2p]}$); $p = 2p$;
15:      $\bar{\theta}^{[1\cdots p]}$ = resampleParticles( $\theta^{[1\cdots p]}, \bar{w}^{[1\cdots p]}$ );
16:      $\overline{\mathcal{GM}}_\theta$ = fitGM( $\bar{\theta}^{[1\cdots p]}$ ); % fit GM on training data
17:      $\mathcal{GM}_\theta \leftarrow \overline{\mathcal{GM}}_\theta$; % becomes prior for the next cycle
18:    notifyRobotToSns($r_{id}$, $\mathcal{GM}_\theta$); % sense for next cycle weight $w^{[i]}$ of $\theta^{[i]}$ (line 6). In the particle filtering context, $w^{[i]}$ represents survival of $\theta^{[i]}$ as per the changed underlying covariance structure. Intuitively, we adapt $\mathcal{GM}_\theta$ to an optimal $\overline{\mathcal{GM}}_\theta^*$ such that the probability of a sample from $\overline{\mathcal{GM}}_\theta^*$ is equal to the posterior probability (conditioning on observations) and the posterior is uniform:

$$\mathcal{H}(\overline{\mathcal{GM}}_\theta^* \mid y) = \mathcal{H}(\overline{\mathcal{GM}}_\theta^*) = \lim_{p \to \infty} -\log(1/p) \quad (4)$$

$$\overline{\mathcal{GM}}_\theta^* = \operatorname{argmin}_{\overline{\mathcal{GM}}_\theta} \mathcal{H}\left(\overline{\mathcal{GM}}_\theta|y\right). \quad (5)$$

Therefore optimality of $\mathcal{GM}_\theta$ can be evaluated by comparison of $\mathcal{H}(\mathcal{GM}_\theta \mid y)$ and $\mathcal{H}(\overline{\mathcal{GM}}_\theta^* \mid y)$. We now present a tractable approximation for $\mathcal{H}(\mathcal{GM}_\theta \mid y)$.

**Lemma 1.** *The conditional entropy on belief $\mathcal{GM}_\theta$ given observations $y$ across $X$ can be written:*

$$\mathcal{H}(\mathcal{GM}_\theta|y) = \lim_{p \to \infty} \log(1/p) - \sum_{i=1}^{p} \bar{w}^{[i]} \log(\bar{w}^{[i]} P(\theta^{[i]})), \quad (6)$$

*where $\bar{w}^{[i]}$ is normalized weight of $\theta^{[i]}$ (line 6, 7).*

*Proof Sketch*[2]: Conditional entropy $\mathcal{H}(\mathcal{GM}_\theta|y)$ is

$$\mathcal{H}(\mathcal{GM}_\theta \mid y) = -\int P(\theta|y) \log P(\theta|y) d\theta$$

$$= -\int \frac{P(y|\theta)P(\theta)}{\int P(y|\theta)P(\theta)d\theta} \log\left(\frac{P(y|\theta)P(\theta)}{\int P(y|\theta)P(\theta)d\theta}\right) d\theta$$

$$= \lim_{p \to \infty} -\sum_{i=1}^{p} \frac{P(y|\theta^{[i]})P(\theta^{[i]})}{\sum_{i=1}^{p} P(y|\theta^{[i]})P(\theta^{[i]})\Delta\theta^{[i]}}$$

$$\log\left(\frac{P(y|\theta^{[i]})P(\theta^{[i]})}{\sum_{i=1}^{p} P(y|\theta^{[i]})P(\theta^{[i]})\Delta\theta^{[i]}}\right) \Delta\theta^{[i]}$$

$$= \lim_{p \to \infty} -\frac{1}{p} \sum_{i=1}^{p} \frac{P(y|\theta^{[i]})}{\frac{1}{p}\sum_{j=1}^{p} P(y|\theta^{[j]})} \log\left(\frac{P(y|\theta^{[i]})P(\theta^{[i]})}{\frac{1}{p}\sum_{j=1}^{p} P(y|\theta^{[j]})}\right)$$

*where the integrals are represented as the Riemann integrals in terms of Riemann sum of $p \to \infty$ no. of terms, and then $P(\theta^{[i]}) \Delta \theta^{[i]}$ is substituted with $1/p$ since only one sample*

[2]See full proof in the extended technical report [19].

$\theta^{[i]} \sim P(\theta)$ *falls in the corresponding $i_{th}$ bin [4, page 120, Eq. 2.241]. Further expanding the expression, we get (6).* ∎

In Lemma 1, the discrete approximation of entropy would be poor if the number of particles $p$ were small. However, in this work, we need the difference of $\mathcal{H}(\mathcal{GM}_\theta \mid y)$ and $\mathcal{H}(\overline{\mathcal{GM}}_\theta^* \mid y)$ to evaluate *percentage of effective particles epp* from $\mathcal{GM}_\theta$ as the direct measure of optimality of $\mathcal{GM}_\theta$ (line 10); both entropy terms are approximately computed using same no. of particles $p$ (line 8, 9). Since $epp$ is compared only against two tuning parameters $opp$ and $spp$ (as we discuss next), this approximation effect can be further nullified with the appropriate tuning.

If $epp$ is more than *optimum particles percentage opp* ($0 \leq opp \leq 100$), the adaptation of the belief is not required (line 11). Otherwise, as part of the belief adaptation procedure at line 12, it is checked if $epp$ is less than stable particles percentage $spp$ ($0 \leq spp \leq opp$). If so, it indicates that $\mathcal{GM}_\theta$ degenerates in modeling the observed dynamics $y$. For eliminating degeneracy, MCMC sampling is performed to sample new particles directly from the posterior $P(\theta|y)$ at line 13 (discussed in details next). For an optimal adaptation of the belief, one can have $opp = 100, spp = 100$. In such case, unless $\mathcal{GM}_\theta$ is the optimal $\overline{\mathcal{GM}}_\theta^*$, belief would be adapted in every sensing cycle; and MCMC sampling would also be performed in each adaptation. Clearly the unconditional adaptation and MCMC sampling would increase computational cost. Considering this direct trade off between adaptation optimality and computational cost, it is recommended to tune $opp$ and $spp$ as per the computational capacity of the server and adaptation accuracy standards. In practice, tuning $opp \in [70, 100], spp \in [20, 50]$ should be efficient.

MCMC sampling has been a successful technique for the general problem of sampling from a posterior distribution. From simulations on our problem, we found Metropolis-Hastings (MH) algorithm [23] more efficient compared to Gibbs sampling [8] (issue of highly correlated samples) and Slice sampling [38] (slow adaptation of slice size). MCMC using Hamiltonian Dynamics [39] is also suitable for the problem though we discuss only MCMC-MH herein (see [4, page 537] for an overview on MCMC). The Markov chain of $\theta$ samples can be started with the highest weighted particle, i.e. $\theta^{(0)} = \theta^{[i^*]}$; $i^* = \operatorname{argmax}_i w^{[i]}$. The proposal distribution $q(\theta|\theta^{(i)})$ can be expressed: $\theta \sim q(\theta|\theta^{(i)}) = \mathcal{N}(\theta|\theta^{(i)}, \Sigma_\theta)$, with the $i^{th}$ Markov chain sample $\theta^{(i)}$ as mean and $\Sigma_\theta$ as the positive semi-definite covariance. The acceptance probability of $\theta$ as $(i+1)^{th}$ sample of the Markov chain is:

$$A(\theta, \theta^{(i)}) = min\left(1, P(y|\theta)q(\theta^{(i)}|\theta)/P(y|\theta^{(i)})q(\theta|\theta^{(i)})\right).$$

MCMC samples $\theta^{(1\cdots p)}$ correspond to particles $\theta^{[p+1\cdots 2p]}$ and the weights $w^{[p+1\cdots 2p]}$. Particles from the prior belief $\mathcal{GM}_\theta$ are kept for resilience to short-term changes in the covariance structure.

As the last step of belief adaptation, the samples $\theta^{[1\cdots p]}$ are resampled using the normalized weights $w^{[1\cdots p]}$ to obtain $\bar{\theta}^{[1\cdots p]}$ (line 15). Then $\overline{\mathcal{GM}}_\theta$ is fit on $\bar{\theta}^{[1\cdots p]}$ (line 16) with *Ex-*

*pectation Maximization* technique [4, page 435]. This adapted $\overline{\mathcal{GM}_\theta}$ is the prior $\mathcal{GM}_\theta$ for the next sensing cycle (line 18).

**Remark 1.** *The proposed belief adaptation technique in Algorithm 1 at line 6–16 solves Problem 4 optimally for $p, k \to \infty$, $spp, opp = 100$. The solution is a low cost approximation with $spp \leq opp < 100$ and finite $p \gg k > 1$.*

*1) Resilience to over convergence:* While resampling is used in standard particle filtering to remove degeneracy of particles, we have already removed degeneracy with MCMC sampling. In this algorithm, resampling is only for shifting the concentration of the non-degenerate set of particles in the hyper-parameters space as per the likelihood weights. In standard particle filtering, in case of high degeneracy, resampled particles are over concentrated in the continuous space (issue of identical particles). Fitting a GM on over concentrated particles would lead to over convergence. However over convergence will not occur in our approach. Regularization in fitting the GM further helps.

*2) Computational cost:* Each GM adaptation step costs $O(p(k + n^3))$ time with $O(n^3)$ for evaluating likelihood $P(\boldsymbol{y}|\boldsymbol{\theta}^{[i]})$ and $O(pk)$ for fitting a $k$-component GM on $p$ samples; $p$ need not be large since dimensionality of the hyper-parameters space is four ($p = 10^3$ for simulations in Sec. V).

*3) Joint sampling from temporally dynamic belief:* The belief is adapted based on the observations of a single robot (line 5) and the same robot samples $\boldsymbol{\theta}$ from that adapted belief (line 18). Thus, one may ask why it is claimed that multiple robots sample from a single GM-based model belief (Problem 3) that is adapted on observations from multiple robots (Problem 4). The reason is that some particles learned from the observations of one robot are expected to survive in the next few GM adaptations. Thus, the GM would represent the belief on the underlying covariance structure of the dynamics sensed by all robots in the recent past. The asynchronous sampling can be interpreted as joint sampling from a GM that is temporally dynamic in the hyper-parameter space.

One can alternatively resort to the synchronized version of Algorithm 1, in which the server waits for all robots to complete sensing before adapting the belief with the joint data. Then, all robots would sample a belief point from the same adapted belief for the decentralized entropy optimizations.

*4) Maximizing joint entropy:* In Algorithm 1, the belief is adapted to $\overline{\mathcal{GM}_\theta}$ such that $\mathcal{H}(\overline{\mathcal{GM}_\theta}|\boldsymbol{y})$ is minimized. Since the adapted belief $\overline{\mathcal{GM}_\theta}$ becomes prior $\mathcal{GM}_\theta$ in the next sensing cycle, it corresponds to maximization of $\mathcal{H}(\mathcal{GM}_\theta)$. The conditional entropy on the dynamics distribution observed by $r$ robots, $\mathcal{H}(\mathcal{Y}^{(1\cdots r)}|\mathcal{GM}_\theta)$, is maximized by decentralized optimizations. The two are related as in Lemma 2.

**Lemma 2.** *Joint entropy $\mathcal{H}(\mathcal{Y}^{(1\cdots r)}, \mathcal{GM}_\theta)$ is the sum of conditional entropy $\mathcal{H}(\mathcal{Y}^{(1\cdots r)}|\mathcal{GM}_\theta)$ and entropy $\mathcal{H}(\mathcal{GM}_\theta)$.*[3]

Therefore, Algorithm 1 maximizes joint entropy $\mathcal{H}(\mathcal{Y}^{(1\cdots r)}, \mathcal{GM}_\theta)$ through the separable maximization of $\mathcal{H}(\mathcal{GM}_\theta)$ and $\mathcal{H}(\mathcal{Y}^{(1\cdots r)}|\mathcal{GM}_\theta)$.

[3]proof for Lemma 2 follows from *chain rules for entropy* [13, page 22].

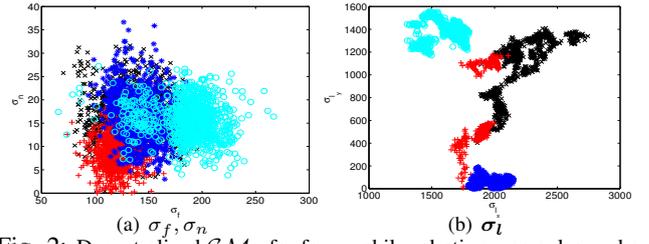

Fig. 2: Decentralized $\mathcal{GM}_\theta$ for four mobile robotic sensors, learned on indoor temperature dynamics, are represented with four different colors. Each GM in the 4-D hyper-parameters space is presented by plotting samples in the corresponding two 2-D spaces separately. The overlap and proximity between the samples from the four distributions demonstrate consensus on model belief amongst the mobile robotic sensors in case of pure decentralized extension (SDE) of the proposed Algorithm 1.

*A. Note on purely decentralized extensions*

In a purely decentralized extension of Algorithm 1, there is no central server; instead each robot locally adapts its belief while communicating only with other robots $\{1\cdots\bar{r}\}$ in the neighborhood. For a belief consensus among robots, we propose: 1) *State Distribution Exchange* (SDE); or 2) *Observation Distribution Exchange* (ODE). In either of the two, there is a small extension at line 6. In SDE, for a robot, $\boldsymbol{\theta}$ is sampled from its own GM and also its neighboring robots': $\boldsymbol{\theta}^{[1\cdots p]} \sim \mathcal{GM}_\theta^{(1\cdots\bar{r})}$ (see Fig. 2 for a demonstration on the effectiveness of this consensus approach). In ODE, given a belief, observations from different robots are assumed conditionally independent. Thus, for a robot, the weight of a sample at line 6 is product of likelihoods of observations sensed by that robot and each of its neighbors: $w^{[i]} = P(\boldsymbol{y}^{(1)}|\boldsymbol{X}^{(1)}, \boldsymbol{\theta}^{[i]})\cdots P(\boldsymbol{y}^{(\bar{r})}|\boldsymbol{X}^{\bar{r}}, \boldsymbol{\theta}^{[i]})$. While the theoretical analysis presented would not be entirely relevant for these purely decentralized extensions, promising empirical results for both extensions are presented in Sec. V.

## IV. Decentralized Optimization of Sensing

In Problem 3, the optimization of sensing locations is decentralized across robots. However, even the decentralized optimization is NP-hard. Now we propose an approximate solution for the decentralized optimization of sensing locations for a single robot, and also address motion planning for visiting the optimized locations.

**Problem 5** (Decentralized Sensing Optimization). *For a hyper-parameters set $\boldsymbol{\theta}$,*
1) *optimize locations $\boldsymbol{X}_* = \{\boldsymbol{x}_*^{(1)}, \cdots, \boldsymbol{x}_*^{(n)} \in \boldsymbol{R}\}$ s.t. $\boldsymbol{X}_* = \operatorname{argmax}_{\boldsymbol{X}:\boldsymbol{X}\subset\boldsymbol{R}} \mathcal{H}(\mathcal{Y}|\boldsymbol{\theta})$ (corresponds to step 2 in Problem 3);*
2) *find a short path $\mathcal{P}(\boldsymbol{X}_*)$ for visiting $\boldsymbol{X}_*$;*
3) *observe $\boldsymbol{y}_* \in \mathbb{R}^n$ across locations $\boldsymbol{X}_*$ along $\mathcal{P}(\boldsymbol{X}_*)$ while optimizing online an obstacle free trajectory $\mathcal{T}(\boldsymbol{x}_*^{(i)}, \boldsymbol{x}_*^{(j)})$ from location $\boldsymbol{x}_*^{(i)}$ to $\boldsymbol{x}_*^{(j)}$.*

In Problem 5, entropy maximization $\operatorname{argmax}_{\boldsymbol{X}} \mathcal{H}(\mathcal{Y}|\boldsymbol{\theta})$ is NP-hard even for the case of selecting a subset of locations from a larger set [30, 33]. Since entropy is a submodular function, the greedy optimization of $n$ spatial locations gives a near-optimal solution $\boldsymbol{X}_s = \{\boldsymbol{x}_s^{(1)}, \boldsymbol{x}_s^{(2)}, \cdots, \boldsymbol{x}_s^{(n)}\}$ (subscript $s$ denotes exploiting submodularity) with entropy value

$\mathcal{H}(\mathcal{Y}_s|\boldsymbol{\theta})$ that is at least $\left(1-\left(\frac{n-1}{n}\right)^n\right)$ of the entropy value $\mathcal{H}(\mathcal{Y}_*|\boldsymbol{\theta})$ for the optimal $\boldsymbol{X}_*$, where $\left(1-\left(\frac{n-1}{n}\right)^n\right)$ converges to $\frac{e-1}{e}$ as $n \to \infty$ [32, 33, 40]. The greedy optimization of $i^{th}$ location can be expressed:

$$\boldsymbol{x}_s^{(i)} = \mathrm{argmax}_{\boldsymbol{x}:\boldsymbol{x}\in\boldsymbol{R}}\mathcal{H}\left(\mathcal{Y}_{\boldsymbol{x}}|\mathcal{Y}_{\boldsymbol{x}_s^{(1\cdots i-1)}}, \boldsymbol{\theta}\right),$$

where $\mathcal{Y}_{\boldsymbol{x}}$, $\mathcal{Y}_{\boldsymbol{x}_s^{(1\cdots i-1)}}$ represent Gaussian distributions across location $\boldsymbol{x}$ and locations set $\boldsymbol{x}_s^{(1\cdots i-1)}$ respectively. Since gradient optimization of the entropy function $\mathcal{H}(\mathcal{Y}_{\boldsymbol{x}}|\mathcal{Y}_{\boldsymbol{x}_s^{(1\cdots i-1)}}, \boldsymbol{\theta})$ in the continuous space would be sensitive to local optima (conditional entropy is multi-modal and so is predictive variance), we propose drawing $p$ MCMC samples $\boldsymbol{x}^{[1\cdots p]}$ of an informative greedy location using the Metropolis Hastings algorithm by defining the likelihood of a location $\boldsymbol{x}$:

$$P(\boldsymbol{x}) = \begin{cases} \exp\left(-1/\mathcal{H}(\mathcal{Y}_{\boldsymbol{x}}|\mathcal{Y}_{\boldsymbol{x}_s^{(1\cdots i-1)}}, \boldsymbol{\theta})^{s_c}\right), & \text{if } \boldsymbol{x} \in \boldsymbol{R} \\ 0, & \text{otherwise} \end{cases}$$

with $s_c$ tuned to amplify the variation of entropy ($s_c=150$ for all simulations). The proposal distribution function for drawing $(j+1)^{th}$ sample in the Markov chain is $q(\boldsymbol{x}|\boldsymbol{x}^{[j]}) = \mathcal{N}(\boldsymbol{x}|\boldsymbol{x}^{[j]}, \boldsymbol{\Sigma}_{\boldsymbol{x}})$ and the acceptance function is

$$A(\boldsymbol{x}, \boldsymbol{x}^{[j]}) = min\left(1, P(\boldsymbol{x})q(\boldsymbol{x}^{[j]}|\boldsymbol{x})/P(\boldsymbol{x}^{[j]})q(\boldsymbol{x}|\boldsymbol{x}^{[j]})\right).$$

Intuitively, greedy optimization of a location in the continuous space using MCMC sampling would give a number of sample locations with equally high likelihood. Thus we get a belief on an informative region in terms of MCMC samples $\boldsymbol{x}^{[1\cdots p]}$. Resampling from $\boldsymbol{x}^{[1\cdots p]}$ with normalized likelihood weights $\bar{w}_{\boldsymbol{x}}^{[1\cdots p]}$ ($w_{\boldsymbol{x}}^{[i]} = P(\boldsymbol{x}^{[i]})$, $\bar{w}_{\boldsymbol{x}}^{[i]} = w_{\boldsymbol{x}}^{[i]}/\sum_{j=1}^p w_{\boldsymbol{x}}^{[j]}$), $\bar{\boldsymbol{x}}^{[1\cdots p]}$ is obtained. Then we fit a GM distribution $\mathcal{GM}_s$ on $\bar{\boldsymbol{x}}^{[1\cdots p]}$ as a continuous parametric representation of the belief on the informative region. Theoretically, the greedy optimization of $n_r$ informative regions can be expressed:

$$\mathcal{GM}_s^{(i)} = \mathrm{argmax}_{\mathcal{GM}_s}\mathcal{H}(\mathcal{Y}_{\mathcal{GM}_s}|\mathcal{Y}_{\mathcal{GM}_s^{(1\cdots i-1)}}, \boldsymbol{\theta})$$

where $\mathcal{Y}_{\mathcal{GM}_s}$ is joint Gaussian distribution across $\mathcal{GM}_s$. Assuming use of mobile sensors (point sensors), it would not be physically possible to sense the dynamics across an entire spatial region and it is also intractable to evaluate the conditional entropy on joint Gaussian distribution across a GM. Therefore, while the above expressions are relevant theoretically, we must approximate each $\mathcal{GM}_s$ with $n_o$ samples for both physical sensing and conditional entropy evaluation (in practice, $n_o \ll p$):

$$\lim_{n_o \to \infty} \mathcal{GM}_s^{(i)} = \mathrm{argmax}_{\mathcal{GM}_s}\mathcal{H}(\mathcal{Y}_{\mathcal{GM}_s}|\mathcal{Y}_{\boldsymbol{X}_o^{(1\cdots i-1)}}, \boldsymbol{\theta})$$

where $\boldsymbol{X}_o^{(i)} \subset \boldsymbol{R}$ represents $n_o$ location samples from the corresponding $\mathcal{GM}_s^{(i)}$. Our simulations suggest that good sensing results can be obtained with $n_o$ as small as 2.

The samples for physical sensing can be different from the ones for conditional entropy approximation so that the locations for sensing can be sampled dynamically from the corresponding GM-based belief of an informative region; in such case, no. of samples from a region for physical sensing is denoted as $n_p$ ($n = n_p n_r$). This allows the flexibility to choose sensing locations which are the product of the robots' online, distributed computation of collision-free trajectories to the informative regions such as those that may be generated from Velocity Obstacles [18].

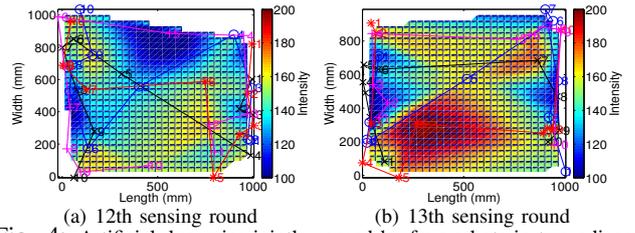

(a) 12th sensing round  (b) 13th sensing round

Fig. 4: Artificial dynamics jointly sensed by four robots in two adjacent rounds. Paths for robots are plotted in different colors with markers representing sensing locations annotated with visiting order. The figure shows that robots complement sensing efforts; sharp temporal changes are also apparent.

The cost for optimizing $n_r$ informative regions ($n_p = n_o$) is $O(n^3 n_r + pn^2 n_r + pn_r k)$ with $p$ small in practice since the samples space is 2-D ($p = 10^3$ in all simulations).

The complete procedure for solving Problem 5 is:

**Procedure 1** (Informative Decentralized Sensing).
1) *learn informative regions:* $\{\mathcal{GM}_s^{(1)}, \cdots, \mathcal{GM}_s^{(n_r)}\}$ *using the proposed MCMC sampling technique;*
2) *sample $n_p$ locations from each region to obtain $n$ sensing locations $\boldsymbol{X}_s$ ($n_p * n_r = n$);*
3) *optimize path $\mathcal{P}(\boldsymbol{X}_s)$ as a Traveling Salesman Problem (TSP) [6, 10, 15];*
4) *make observations $\boldsymbol{y}_s$ across $\boldsymbol{X}_s$ along $\mathcal{P}(\boldsymbol{X}_s)$ while optimizing $\mathcal{T}(\boldsymbol{x}^{(i)}, \boldsymbol{x}^{(j)})$ online using the* Velocity Obstacles *algorithm.*

**Remark 2.** *For $p, n_o, k \to \infty, n_p = 1, n = n_r$, Procedure 1 solves Problem 5 near optimally while assuming optimal solutions from the TSP algorithm and the Velocity Obstacle algorithm. For finite $n_o, p, k$ s.t. $1 < n_o \leq k \ll p$, the solution is an approximation.*

**Remark 3.** *For optimal solutions from Algorithm 1 and Procedure 1, Problem 1 is optimally solved.*

A reasonable approximation to optimum can be obtained with low computational cost (as demonstrated next in Sec. V).

## V. EMPIRICAL EVALUATION

We evaluated the proposed algorithm in MATLAB simulations using one artificial and four real datasets of previous static sensor deployments. For the real datasets, the sensory function value in the continuous space is interpolated.

*1) Artificial:* A set of 220 temporally moving sinusoidal and gamma multi-scale fields create a spatio-temporal stochastic sensory function. The fields' temporal movement leads to the change of the underlying covariance structure for rigorous testing of the belief adaptation technique. The complex multi-scale dynamics help in benchmark validation of the proposed multi-scale decentralized sensing technique (see Fig. 4).

*2) Temperature:* On 46 wireless sensor motes in an indoor region of $45\times40$ m$^2$ (Intel Berkeley lab in Feb, 2004), temperature was sensed every 30s (sensing simulated every second) for 5 days, from 7 AM–7 PM every day [33].

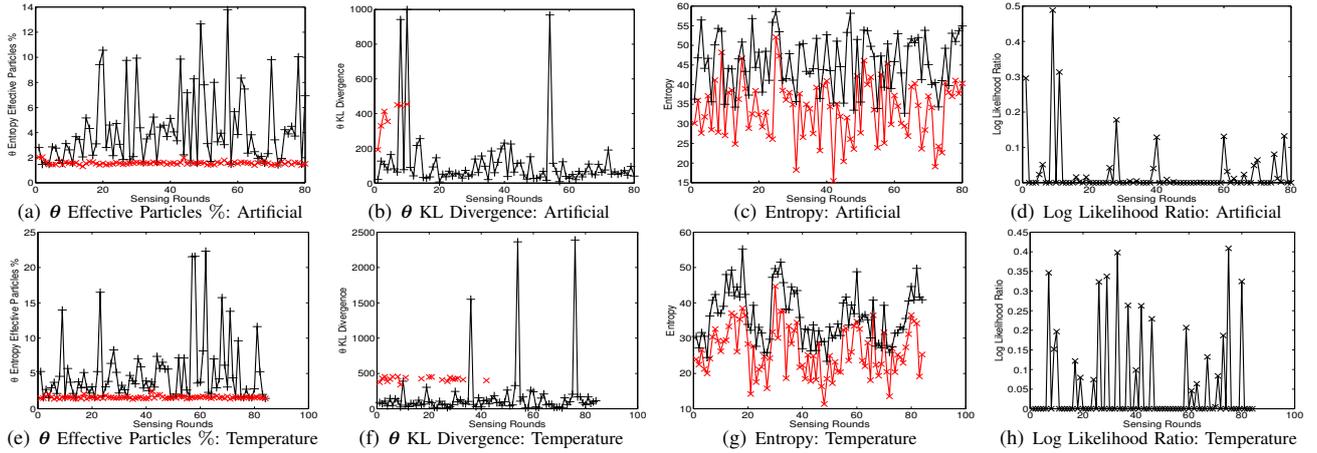

Fig. 3: Baseline evaluation of the proposed system solution on artificial (a-d) and indoor temperature (e-h) datasets. Red 'x's represent random selection.

*3) Sunlight:* Forest canopy understory sunlight intensity images collected every 10m (sensing simulated every 100s) from 8 AM–8 PM covering approx. $6\times4$ m$^2$ [44].

*4) Windspeed:* Daily average wind speed sensed in 1961-78 at 12 stations in Ireland [21]; daily sample sensing simulated every second in this setup.

*5) Precipitation:* Daily data collected for rainy days in 1949–94 across 167 regions in Washington and Oregon states; log of daily mean sensed every second in this setup [33].

### A. Simulated Sensing System Setup

We run simulations on a 2.1 GHz Intel i3 laptop with 4GB of RAM using the MATLAB 2013b parallel computing toolbox. We run a simulation on each dataset for approximately 5000s such that it covers a temporal change of: indoor temperature from day 1 to 4 (12 hours every day); windspeed from 1961–1974; sunlight from 8:00 AM–4:20 PM; and precipitation from the 1st to the 5000th rainy day. For each dataset, the spatial space is scaled to $1000\times1000$ square units.

For the belief adaptation algorithm, $spp = 20$, $opp = 80$. For some experiments, communication failure with probability 0.3 is simulated between the server and all robots. For SDE and ODE, communication range radius is 300 units. Unless noted otherwise, four robots ($r = 4$) are used for sensing where each robot optimizes on five informative spatial regions ($n_r = 5$) and two sensing locations ($n_o = n_p = 2$) per region in each sensing cycle. A robot, with body size and sensing range of 10 units radius, plans a short path for visiting the sensing locations as the TSP. The trajectory is optimized online using Velocity Obstacles where another robot is a potential obstacle if within 600 units radius. For experiments analyzing the effect of localization error, the observation is sensed at a location $\boldsymbol{x} \sim \mathcal{N}(\boldsymbol{x_c}, diag([10, 10]))$ where $\boldsymbol{x_c}$ is known as the current location to the robot. For robot motion, max. acceleration is $[300, 300]$ units/s$^2$ (in 0.1s); min. and max. values for velocity are $[-30, -30]$, $[30, 30]$ units/s respectively.

### B. Evaluation after each sensing round

*1) Evaluating GM-based belief adaptation:* In Algorithm 1, belief on the covariance structure is adapted after every sensing round as a GM in the hyper-parameter space of the GP. The adapted hyper-parameters distribution is evaluated, while using a true distribution as a baseline, in terms of: a) *percentage of effective particles* (%EP) from the adapted distribution that contributes entropy; b) *KL-Divergence* (KL-D) between the adapted and true distribution. In Algorithm 1, the belief is initialized randomly. The two metrics are also computed on the initial distribution for a comparison against adaptation. In Fig. 3(a), 3(e), the %EP for every sensing round is plotted for both distributions (red color represents the initial distribution). As expected, the %EP remains close to zero for the initial distribution, but increases to 25% for the adapted distribution. Zigzaging in Fig. 3(a), 3(e) indicates the GM adapted on the dynamics observed by one robot can be more representative than that of others. The consistent zigzag also indicates that the joint effort of the robots ensures an accurate belief adaptation.

In Fig. 3(b), 3(f), the black curve represents KL-D between the adapted and true distribution; the red curve represents KL-D between the initial and true distribution (missing red 'x's represent infinite KL-D). KL-D value represents the overlap between distributions (inverse relation). Since the initial distribution is static, the high increase in KL-D value along the red curve indicates that true distribution, i.e. the phenomenon's underlying covariance structure, has changed significantly. These temporal changes in the covariance structure of the temperature and artificial dynamics allow us to analyze the effectiveness of Algorithm 1 in adapting the belief since there is reasonable overlap between the adapted and true distribution across all sensing rounds (low KL-D along black curves).

*2) Evaluating representativeness in a single round:* We evaluate representativeness of locations sensed by a robot in terms of: 1) entropy on phenomenon dynamics distribution across the sensed locations; and 2) log likelihood of the unobserved dynamics conditioned upon the observations across the sensed locations. The informativeness of sensed locations is evaluated against the random selection of locations using both metrics. Fig. 3(c), 3(g) show that entropy of the dynamics distribution across the sensed locations is indeed higher than random case (red color for random). In Fig. 3(d), 3(h), ratio of log likelihood with sensed locations and random selection of locations is plotted. Since log likelihood ratio is very close to zero for most of the rounds (and always less than 1) in the

TABLE I: KL-Divergence on Observed Spatio-Temporal Datasets.
(a) Variation across robots. RobotID (RId) "Jt" represents joint data from all robots.
(b) Varying no. of robots for Proposed-PDE-ODE ($n_r = 10, n_o = 1$)

| RId | Art. | Temp. | Wind | Sun. | Prec. |
|---|---|---|---|---|---|
| Jt | **0.0013** | **0.0132** | **0.0008** | **0.0058** | **0.0009** |
| 1 | 0.0243 | 0.0405 | 0.0209 | 0.0220 | 0.0090 |
| 2 | 0.0297 | 0.0205 | 0.0285 | 0.0224 | 0.0156 |
| 3 | 0.0236 | 0.0182 | 0.0195 | 0.0118 | 0.0100 |
| 4 | 0.0234 | 0.0156 | 0.0215 | 0.0197 | 0.0127 |

| Data | $r=2$ | $r=4$ | $r=8$ | $r=12$ |
|---|---|---|---|---|
| Art. | 0.0018 | 0.0010 | **0.0006** | **0.0005** |
| Temp. | 0.0033 | 0.0019 | 0.0017 | **0.0011** |
| Wind | 0.0080 | 0.0010 | **0.0005** | **0.0004** |
| Sun. | 0.0088 | 0.0016 | **0.0008** | **0.0007** |
| Prec. | 0.0028 | 0.0009 | **0.0005** | **0.0005** |

(c) Competitive approaches ($r=4, n_r=10, n_o=1$)

| Approach | Art. | Temp. | Wind | Sun. | Prec. |
|---|---|---|---|---|---|
| Proposed | **0.0011** | **0.0022** | **0.0008** | **0.0013** | **0.0008** |
| Proposed-PDE-SDE | 0.0014 | 0.0096 | **0.0009** | 0.0024 | **0.0009** |
| Proposed-PDE-ODE | **0.0010** | **0.0019** | 0.0010 | 0.0016 | **0.0009** |
| SA-S (Centralized) | 0.0016 | 0.0102 | 0.0023 | 0.0032 | 0.0817 |
| SA-GMTA (Centralized) | 0.0075 | 0.0277 | 0.0043 | 0.0062 | 0.0078 |
| RIG (Decentralized) | 0.0036 | 0.0080 | 0.0023 | 0.0038 | 0.0027 |
| MCES (Decentralized) | 0.1373 | 1.1509 | 0.1631 | 0.1696 | 0.1954 |
| Proposed-communication failure | 0.0012 | 0.0052 | 0.0013 | 0.0016 | 0.0011 |
| Proposed-localization error | 0.0015 | **0.0018** | **0.0008** | **0.0012** | **0.0009** |

plots, it means that the observations across sensed locations predict the unobserved dynamics with much higher likelihood than observations across a random selection of locations.

### C. Evaluation on complete datasets

Now we present a unified evaluation of the complete spatio-temporal dataset obtained from multiple sensing rounds executed in the total 5000s duration. For evaluation of a sensed spatio-temporal dataset, the corresponding observed dynamics distribution is learned as a GM in 4-D space (2-D for spatial space, 1-D for time, 1-D for observations). This observed distribution is evaluated in terms of KL-Divergence between the observed and a true phenomenon dynamics distribution (not to be confused with a true hyper-parameter distribution used for the baseline evaluation of belief adaptation). Lower values of KL-D indicate the sensed spatio-temporal dataset better represents the true phenomenon.

KL-D comparison on datasets sensed by each of the four robots is in Table I(a) (robot id "Jt" represents the sensing dataset of all robots). The table shows that some robots perform better than others in representing the true distribution. Since KL-D is significantly lower for the joint case than the individual cases, one can infer that the robots' observed distributions are complementary, and therefore give a highly representative joint observed distribution.

From the analysis of the effect on KL-D of varying the number of robots $r$ in Table I(b), we observe that the decrease in KL-D value with an increase in $r$ becomes negligible eventually. This indicates that a small team of robots (about 8) is sufficient for optimal sensing of the phenomenon dynamics.

Table I(c) presents KL-D of the joint dataset using our proposed solution, its decentralized extensions SDE and ODE, and some competitive recent approaches. When comparing results for variants of our proposed approach, the original ("Proposed") and decentralized extensions (with suffix "PDE") perform equally well except in the temperature dataset. One reason for this may be that entropy on the boundary of the indoor region is higher, which could lead the robots to sense the dynamics more along the boundaries. In such case, robots may not be able to communicate with each other (communication radius is 1/3 of length of indoor hall). Communication failure amongst robots apparently leads to the disadvantage on performance with SDE.

Singh et al. [45] propose to extend a single mobile sensing optimization to the multiple case in a centralized manner using their generic Sequential Allocation (SA) algorithm. Since a single information function can be used in the centralized optimization, entropy is evaluated considering both cases: (SA-S) using a single sample from the covariance belief; or (SA-GMTA) using multiple samples from the belief as per the approximation in (2). While SA-GMTA does not perform as well, SA-S performs comparatively with our proposed approach for some datasets (mind that information optimization is decentralized in our approach and its variants). For comparative evaluation of our approximate algorithm for optimizing information for a single robot (Sec. IV), we evaluate it against the Rapidly Exploring Information Gathering (RIG) algorithm [25]. Table I(c) shows that our approach outperforms RIG. Another competitive class of works [24, 27, 28] optimize on sensing locations by minimizing conditional entropy on the underlying state (abbreviated MCES). This corresponds to maximizing mutual information between the observed dynamics distribution and the belief distribution on the state, whereas our work maximizes joint entropy on both (Lemma 2). Results for MCES validate the higher comparative applicability of our approach. Furthermore, results from the simulations accounting for the effects of localization and communication errors demonstrate the robustness of the proposed algorithm.

### VI. DISCUSSION

In this work, we have proposed a solution for persistently sensing the spatio-temporal dynamics of a real world stochastic phenomenon using a small team of mobile robots. The approach uses a central server to adapt the belief on the covariance structure of the phenomenon dynamics using asynchronously communicated sensing data from the robots. The belief is adapted using a combined particle filtering and MCMC approach, which ensures accurate belief adapation even when the dynamics change sharply.

We have also proposed a novel MCMC algorithm to approximate optimizing informative regions in continuous space, which is known to be intractable; this is executed in a distributed way, each robot using a belief point sample from the most up-to-date model available. The ability to choose informative regions instead of specific locations provides flexibility towards online optimization of collision-free trajectories to the sensing locations in the informative regions.

With extensive baseline and comparative sensing simulations, we show the proposed approach is effective for monitoring complex phenomena using a small number of robots. We also discuss and present comparable results for a fully decentralized modification to our approach where the belief on covariance structure is also adapted in a decentralized manner.

### VII. ACKNOWLEDGEMENTS

We are very grateful to Prof. Greg Ver Steeg (USC) for discussions on Lemma 1, and to Prof. Amarjeet Singh (IIIT Delhi) for access to the sunlight dataset.